\documentclass[conference]{IEEEtran}
\usepackage{cite}
\usepackage{amsmath,amssymb,amsfonts}
\usepackage{algorithmic}
\usepackage{graphicx}
\usepackage{textcomp}
\usepackage{xcolor}
\usepackage{booktabs}
\usepackage{url}
\usepackage{tikz}
\usepackage{pgfplots}
\usepackage{hyperref}
\usepackage{multirow}
\usepackage{tabularx}
\usepackage{placeins}
\usepackage{array}
\newcolumntype{C}{>{\centering\arraybackslash}X}
\usepgfplotslibrary{groupplots}
\usetikzlibrary{patterns}
\pgfplotsset{compat=1.18}
\usetikzlibrary{shapes.geometric, arrows.meta, positioning, fit, backgrounds, calc}

\definecolor{pubblue}{rgb}{0.00,0.44,0.75}
\definecolor{pubgreen}{rgb}{0.00,0.65,0.31}
\definecolor{pubred}{rgb}{0.85,0.16,0.10}
\definecolor{puborange}{rgb}{1.00,0.60,0.07}

\def\BibTeX{{\rm B\kern-.05em{\sc i\kern-.025em b}\kern-.08em
    T\kern-.1667em\lower.7ex\hbox{E}\kern-.125emX}}
\begin{document}

\title{Beyond the Stitching Assumption: A Unified Framework for Multimodal Synthetic Data Evaluation via Semantic Quantization}

\author{Yefeng Yuan \\
  Santa Clara University\\
  Santa Clara, USA \\
  \texttt{yyuan4@scu.edu} \and
  Zhan Shi \\
  Santa Clara University\\
  Santa Clara, USA \\
  \texttt{ashi2@scu.edu} \and
  Liang Cheng \\
  eBay Inc.\\
  San Jose, USA \\
  \texttt{liacheng@ebay.com} \and
  Yuhong Liu \\
  Santa Clara University\\
  Santa Clara, USA \\
  \texttt{yhliu@scu.edu} \\
  }

\maketitle

\begin{abstract}
Multimodal synthetic datasets combine structured attributes with free text, but are often evaluated separately. Such metrics can remain high after tabular--text pairings are disrupted. We present a projection-based evaluator for tabular--text synthetic data. A fixed sentence encoder maps text to embeddings, \(k\)-means converts them to cluster states, and tabular variables are represented as categorical or quantile-binned states. Real and synthetic contingency tables are compared using Jensen--Shannon divergence (JSD), normalized mutual information (NMI), conditional JSD (cJSD), and joint-state entropy. We also report text-to-attribute (T2A) utility and a holdout-calibrated proximity flag rate (PFR) as a representation-level diagnostic. A text-permutation control preserves both marginal distributions while disrupting their pairing. Experiments on Amazon Reviews, Kiva Loans, and the Employment Scam Aegean Dataset show that modality-specific scores remain high under this control. The projection diagnostics detect disruption when real projected dependence exceeds a permutation baseline, but are less informative for weak or sparse projections. Some conditioned LLM baselines also exhibit stronger measured dependence than the corresponding real-data projections. These results support explicit cross-modal evaluation with permutation baselines and coverage reporting.
\end{abstract}


\section{Introduction}
Recent advances in generative models have made synthetic data a practical tool for data sharing, augmentation, and benchmarking in privacy-sensitive domains~\cite{wef2025synthetic}. Many modern applications, especially in healthcare, finance, and enterprise analytics, involve multimodal records that pair structured attributes with unstructured text. Examples include physiological measurements paired with clinical notes~\cite{johnson2016mimic,huang2019clinicalbert}, loan metadata paired with borrower narratives~\cite{kiva2018kaggle}, job-posting metadata paired with job descriptions~\cite{vidros2017automatic}, and product metadata paired with review text~\cite{hou2024amazon}. In these settings, synthetic data is useful only if it can be evaluated for both semantic consistency and privacy risk.

As generative architectures ranging from Generative Adversarial Networks (GANs) and diffusion models to Large Language Models (LLMs) are applied to multimodal records, evaluation methods have not kept pace with the structure of the generated data. Existing metrics are often modality-specific. Tabular evaluation commonly relies on marginal distribution tests, such as the Kolmogorov--Smirnov test and total variation (TV) distance, together with low-order dependency statistics over structured features, as implemented in widely used synthetic-data evaluation tools such as SDMetrics and Synthcity~\cite{patki2016synthetic,qian2023synthcity}. Text evaluation, by contrast, uses reference-similarity metrics such as BERTScore~\cite{zhang2019bertscore} and distributional metrics such as MAUVE~\cite{pillutla2021mauve}. Because these metrics evaluate text without conditioning on the tabular attributes in the same record, they do not by themselves test row-level cross-modal consistency. As a result, an evaluator may validate each modality separately while missing broken row-level alignment, sensitive cross-modal associations, or synthetic records that remain too close to sensitive originals.

\begin{figure}[t]
    \centering
    \includegraphics[width=\columnwidth]{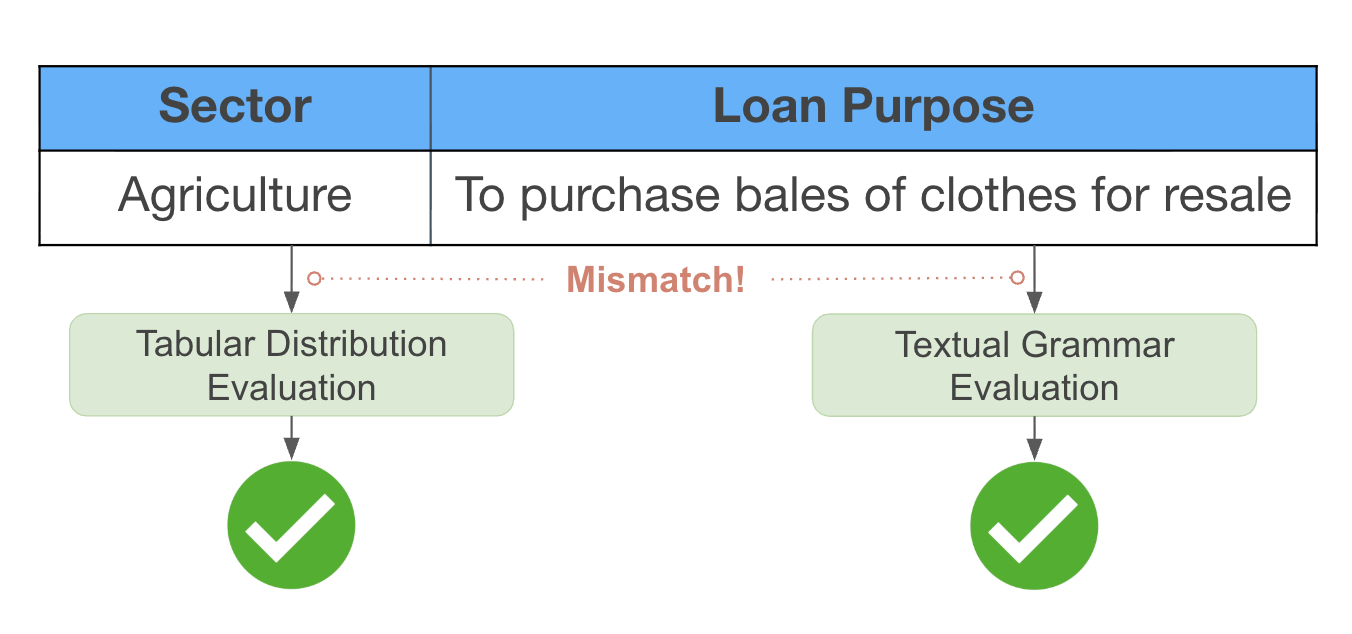}
    \caption{The ``stitching assumption'' in multimodal synthetic data evaluation. Separate tabular and text metrics can assign high scores even when the two components of the same record contradict each other.}
    \label{fig:stitching_assumption}
\end{figure}

This separation leads to what we call the ``stitching assumption'': the belief that strong tabular-only and text-only scores imply a coherent multimodal record. For example, a synthetic loan record may pair a sector of ``Agriculture'' with a fluent loan purpose stating ``to purchase bales of clothes for resale'' (as illustrated in Figure~\ref{fig:stitching_assumption}). The sector distribution may look realistic, and the loan purpose may be fluent, but the paired record is incoherent. In privacy-sensitive settings, separate modality-level checks can also miss full-record proximity to training examples or repeated tabular--text pairings; more broadly, they can miss spurious cross-modal associations that affect downstream validity.

A natural evaluation target is the cross-modal joint structure between tabular states and text semantics. One nonparametric way to compare this structure is to quantize the variables and construct contingency tables over their joint states. However, a full table over all tabular attributes and text-embedding clusters grows with the Cartesian product of the state spaces and quickly becomes sparse. Richer probabilistic models can represent more complex joint structure, but they are often too costly, assumption-dependent, or difficult to use as routine evaluation tools~\cite{silverman2018density,koller2009probabilistic,cooper1990computational}. Our goal is therefore not to recover the full continuous multimodal joint distribution. Instead, we evaluate selected low-dimensional quantized cross-modal projections. Here, a projection denotes a selected view, such as one tabular attribute or a small tabular subset crossed with a text semantic-state variable. These projections are designed to be statistically estimable, interpretable, and useful for diagnosing cross-modal dependence.

\begin{figure*}[t]  
    \centering
    \includegraphics[width=\textwidth]{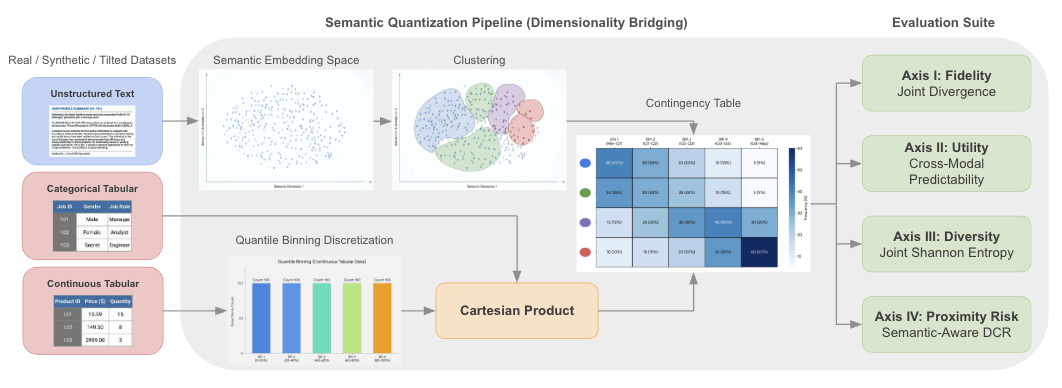}
    \caption{Architectural overview of the proposed framework. Semantic quantization maps text embeddings and selected tabular features into a joint discrete representation, enabling coarse-grained joint evaluation of multimodal synthetic data across fidelity, dependence, utility, diversity, and record proximity.}
    \label{fig:framework}
\end{figure*}

To address this dimensionality bottleneck, this work extends SynEval\footnote{Code is available at \url{https://github.com/privacy-enhancing-technologies/SynEval}.}, our earlier framework for multifaceted synthetic-data evaluation~\cite{yuan2024multi}, to multimodal tabular--text synthetic-data evaluation, as shown in Figure~\ref{fig:framework}. The original SynEval framework focused on decoupled quality, utility, diversity, and privacy axes for LLM-generated synthetic data. In contrast, the present work adds a semantic-quantization layer for evaluating tabular--text association: text sequences are mapped to dense embeddings and grouped into discrete semantic clusters, while tabular features are discretized into categorical or binned states. The resulting contingency tables allow direct empirical comparison within selected cross-modal projections. This design does not eliminate the curse of dimensionality; instead, it turns joint evaluation into an explicit projection-selection problem. The scalable default uses pairwise tabular--text projections, while the same framework supports budgeted higher-order tabular projections crossed with a text semantic-state variable when cell budget and support conditions are satisfied.

Our contributions are threefold:
\begin{enumerate}
    \item \textbf{A controlled test of modality-isolated evaluation.}
    We formalize the ``stitching assumption'' as an evaluation failure mode in which favorable tabular-only and text-only scores are treated as evidence of a coherent multimodal dataset. We introduce a text-permutation control that preserves the marginal distribution of each modality while disrupting their pairing.

    \item \textbf{A projection-based extension of SynEval for tabular--text data.}
    We map text into fixed embedding-cluster states and tabular variables into categorical or quantile-binned states, then compare selected tabular--text contingency tables. The method evaluates tractable low-dimensional projections rather than attempting to estimate the full continuous multimodal joint distribution.

    \item \textbf{A multi-axis evaluation with explicit boundary conditions.}
    We evaluate projected fidelity, measured dependence, predictive utility, joint-state entropy, and representation-level record proximity on three datasets. The experiments also identify conditions under which the diagnostics become less informative, including weak real-data dependence, sparse projected support, and higher-order interactions that are not visible in pairwise projections.
\end{enumerate}

\section{Related Work}
\label{sec:relatedwork}

Synthetic-data research is increasingly motivated by data-access constraints, model development needs, and privacy regulations~\cite{gdpr2016, ccpa2018}. For multimodal records, however, the literature remains fragmented across generation methods, modality-specific evaluation metrics, and separate privacy or diversity diagnostics. We therefore review three strands of related work: generative architectures for tabular and textual modalities, limitations of modality-isolated evaluation, and the dimensionality, diversity, and proximity challenges that arise in joint multimodal evaluation. Cross-modal dependence is well established in multimodal learning; our contribution concerns the narrower problem of evaluating whether synthetic tabular--text datasets preserve it under controlled negative tests and selected quantized projections~\cite{baltrusaitis2018multimodal}.

\subsection{Generative Architectures for Tabular and Textual Modalities}

Structured synthetic data has been generated using statistical models such as Bayesian networks and copulas~\cite{heckerman2008tutorial,frees1998understanding}, as well as neural generators such as Conditional Tabular Generative Adversarial Network (CTGAN), Tabular Variational Autoencoder (TVAE), and diffusion models~\cite{xu2019modeling,kotelnikov2023tabddpm,zhou2025missddim,villaizan2025diffusion}. These methods improve flexibility but remain sensitive to heterogeneous feature types, preprocessing, and data imbalance. Text synthesis has been advanced by LLMs~\cite{brown2020language,touvron2023llama}, including prompt-based and fine-tuned approaches for tabular data~\cite{kim2024epic,borisov2022language,zhao2023tabula}. However, generating tabular attributes and text as a coherent record remains difficult: some methods flatten tables into text or use late fusion rather than explicitly preserving row-level cross-modal dependence~\cite{hegselmann2023tabllm,fang2024large,baltrusaitis2018multimodal,gaw2022multimodal}. This motivates evaluation metrics that measure tabular--text alignment rather than unimodal realism.

\subsection{Limitations of Modality-Isolated Quality Metrics}

Commonly used synthetic-data evaluation toolkits mostly focus on unimodal data types, rather than multimodal combinations. SDMetrics~\cite{patki2016synthetic} evaluates univariate column shapes and selected pairwise trends among structured variables, while Synthcity~\cite{qian2023synthcity} provides broader fidelity, utility, and privacy evaluations for several tabular-data settings. These toolkits do not natively assess whether free-text fields are consistent with the structured attributes in the same record.

In natural-language generation, the Bilingual Evaluation Understudy (BLEU)~\cite{papineni2002bleu} and Recall-Oriented Understudy for Gisting Evaluation (ROUGE)~\cite{lin2004rouge} measure reference overlap, BERTScore~\cite{zhang2019bertscore} measures contextual candidate--reference similarity, and MAUVE~\cite{pillutla2021mauve} compares generated and human-written text distributions. MAUVE also quantizes language-model representations, whereas our method crosses fixed text-embedding clusters with tabular states to evaluate cross-modal joint structure.

Our earlier SynEval framework provided a multifaceted evaluation for LLM-generated synthetic data, including quality, utility, diversity, and privacy diagnostics~\cite{yuan2024multi}. That framework was designed primarily around modality-level and task-level evaluation axes, and it did not explicitly test whether structured attributes and unstructured text remain mutually consistent within the same row. The present work extends SynEval by adding semantic quantization, selected cross-modal contingency tables, the text-permutation control, and projection-level diagnostics for tabular--text alignment.

\subsection{Joint Evaluation Challenges: Scale, Diversity, and Proximity}

One common component of synthetic-data risk evaluation is detecting synthetic records that are unusually close to training records, since such records may indicate memorization or increased disclosure risk. The Distance to Closest Record (DCR) is often used as an empirical nearest-neighbor proximity diagnostic for this purpose~\cite{choi2017generating}. For tabular--text records, however, DCR is not meaningful without an explicit representation and scaling convention. Naively concatenating encoded tabular features with high-dimensional text embeddings can make Euclidean distances dominated by one modality rather than by meaningful full-record similarity~\cite{aggarwal2001surprising}. This motivates calibrated proximity diagnostics that specify the multimodal representation and interpret metrics like DCR as empirical record-proximity screens rather than formal privacy guarantees.

Diversity evaluation raises a different but related joint-structure problem. Generative diversity is often assessed with isolated measures such as tabular support coverage, marginal entropy, or textual vocabulary richness. These measures can miss an analogue of mode collapse in cross-modal pairings: a generator may produce diverse tabular values and diverse text, but combine them through a narrow set of repetitive tabular--text associations. Joint entropy provides a standard information-theoretic measure of spread over joint states~\cite{cover2006elements} and has also been used in subset-selection settings such as sensor placement~\cite{krause2006near}. We therefore use joint entropy as a projection-level diagnostic of diversity over quantized cross-modal states, while interpreting high entropy together with dependence and fidelity metrics rather than as evidence of semantic validity by itself.

These limitations motivate a projection-based evaluator that avoids estimating the full joint distribution. Evaluations should look beyond isolated modality scores, but they must avoid full joint modeling procedures that are too sparse or costly for routine use. This motivates our semantic-quantization approach, which compares selected quantized cross-modal projections as tractable, interpretable diagnostics of multimodal synthetic records.

\section{Methodology}
\label{sec:methodology}

The proposed framework evaluates multimodal synthetic data through selected quantized cross-modal projections and fixed-representation diagnostic probes. Instead of estimating the full continuous multimodal joint distribution, it maps text into discrete embedding-cluster states and tabular variables into categorical or quantile-binned states for projection-level fidelity, dependence, and diversity diagnostics, while also reporting cross-modal utility and calibrated record-proximity diagnostics. All data-dependent evaluation artifacts---including text-cluster quantizers, tabular discretizers, categorical vocabularies, normalization statistics, and the record-proximity scaling parameter \(\lambda^\star\)---are fitted on the real training partition and applied unchanged to holdout and synthetic records. The Sentence-BERT (SBERT) encoder itself is pretrained and kept fixed.

Throughout the paper, \(D_{\mathrm{demo}}\) denotes the fixed 10-example prompt-demonstration set, \(R_{\mathrm{train}}\) and \(R_{\mathrm{holdout}}\) denote the disjoint real training and holdout partitions, \(R=R_{\mathrm{train}}\cup R_{\mathrm{holdout}}\) denotes the real evaluation pool, and \(S\) denotes a synthetic dataset. No demonstration record appears in either \(R_{\mathrm{train}}\) or \(R_{\mathrm{holdout}}\). Unless stated otherwise, projection-level real reference distributions are computed on \(R\) using quantizers fitted on \(R_{\mathrm{train}}\), whereas predictive-probe evaluation and holdout-based proximity-threshold calibration use \(R_{\mathrm{holdout}}\).

\subsection{Semantic Quantization and Projected State Spaces}

Let \(\mathcal{D}=\{(\mathbf{x}^{(i)},\mathbf{t}^{(i)})\}_{i=1}^{N}\) denote a generic multimodal dataset, where \(\mathbf{x}^{(i)}\) contains structured tabular attributes and \(\mathbf{t}^{(i)}\) denotes the predefined text field used in the evaluated projection. Directly estimating the full Cartesian product over all tabular and text states is infeasible: if tabular feature \(j\) has \(|V_j|\) states and text field \(l\) has \(K_l\) semantic clusters, the full joint state space scales as:
\begin{equation}
|\mathcal{C}_{\mathrm{full}}|
=
\left(\prod_{j=1}^{m}|V_j|\right)
\left(\prod_{l=1}^{p}K_l\right).
\label{eq:state_explosion}
\end{equation}

Our framework therefore evaluates selected projections rather than the full multimodal joint distribution. The scalable default is pairwise: one tabular-state variable \(C_{X_j}\) is evaluated against one text semantic-state variable \(C_T\). This setting is used for all real-data projections in the main experiments because it is more tractable and interpretable than higher-order projections. Empirical support is assessed separately for each dataset and semantic resolution \(K\), and sparse projections are explicitly flagged.

\paragraph{Budgeted multivariate projections.}
The pairwise setting is scalable and interpretable, but it can miss dependencies that only appear when several tabular variables are considered jointly. The framework therefore supports budgeted multivariate projections over a user-specified tabular subset \(A\), yielding a joint tabular state \(C_{X_A}\). The multivariate extension is a candidate-projection evaluator rather than an automatic structure-discovery algorithm: higher-order projections must be specified by domain knowledge, a predefined evaluation plan, or a targeted stress test.

A candidate multivariate projection may be screened using the following state-space and real-support conditions:
\begin{equation}
\begin{aligned}
|\mathcal{C}_{X_A}|\,K &\le B_{\max},\\
\operatorname{cov}_{\ge n_{\min}}(A)
&= \sum_{u\in\mathcal U_A} p_R(u),\\
\operatorname{cov}_{\ge n_{\min}}(A)
&\ge \rho,
\end{aligned}
\label{eq:projection_budget}
\end{equation}
where \(p_R(u)=\hat P_R(C_{X_A}=u)\), and
\(\mathcal U_A=\{u:n_R(C_{X_A}=u)\ge n_{\min}\}\) is the set of real-data joint tabular states with at least \(n_{\min}\) records. \(B_{\max}\), \(n_{\min}\), and \(\rho\) are user-specified evaluation-budget parameters rather than learned model parameters. The reported real-data experiments use only the pairwise case \((|A|=1)\); the XOR/parity experiment is a prespecified stress test rather than a projection selected by this screening rule. Here, \(B_{\max}\), \(n_{\min}\), and \(\rho\) are evaluation-budget parameters rather than learned model parameters. \(B_{\max}\) limits the number of cells in the projected contingency table, \(n_{\min}\) defines the minimum real-data support required for a tabular joint state to be treated as stable, and \(\rho\) requires that such stable states cover a sufficient fraction of the real-data mass.

In our implementation, text fields are encoded with a fixed Sentence-BERT encoder and clustered with \(k\)-means fitted on real training embeddings, yielding the text semantic-state variable \(C_T\). Continuous tabular features are discretized using training-split quantile bins, while categorical features retain their observed training-split categories, yielding tabular-state variables \(C_{X_j}\). For any selected projection \((C_X,C_T)\), the empirical real projected distribution is
\begin{equation}
\hat P(C_X=u,C_T=v)
=
\frac{1}{|R|}
\sum_{i\in R}
\mathbb{I}
[
c_X^{(i)}=u
\land
c_T^{(i)}=v
],
\label{eq:joint_pmf}
\end{equation}
and the synthetic projected distribution \(\hat Q\) is defined analogously on \(S\).

\subsection{Axis I: Projection Fidelity via Jensen--Shannon Divergence}

Projection fidelity measures whether the synthetic quantized table matches the real quantized table. We use Jensen--Shannon divergence (JSD), defined in terms of Kullback--Leibler (KL) divergence:
\begin{equation}
\begin{aligned}
\mathrm{JSD}(\hat P\parallel \hat Q)
&=
\tfrac{1}{2}D_{\mathrm{KL}}(\hat P\parallel M)
+
\tfrac{1}{2}D_{\mathrm{KL}}(\hat Q\parallel M),\\
M&=\tfrac{1}{2}(\hat P+\hat Q).
\end{aligned}
\label{eq:jsd}
\end{equation}
With base-2 logarithms, JSD lies in \([0,1]\), and lower values indicate higher projected fidelity. Importantly, JSD is an overall table-fidelity metric: it can increase because of tabular marginal mismatch, text-cluster marginal mismatch, or dependence mismatch. We therefore complement it with dependence-preservation diagnostics. We use the convention \(0\log 0=0\).

\subsection{Axis II: Cross-Modal Dependence via Normalized Mutual Information and Conditional Jensen--Shannon Divergence}

To more directly measure dependence preservation, we compute normalized mutual information (NMI):
\begin{equation}
\mathrm{NMI}(C_X,C_T)
=
\frac{I(C_X;C_T)}
{\sqrt{H(C_X)H(C_T)}}.
\label{eq:nmi}
\end{equation}
We report \(\mathrm{NMI}_{\mathrm{real}}\), \(\mathrm{NMI}_{\mathrm{synth}}\), the signed gap
\(\Delta_{\mathrm{NMI}}=\mathrm{NMI}_{\mathrm{real}}-\mathrm{NMI}_{\mathrm{synth}}\), and the ratio
\(\mathrm{NMI}_{\mathrm{synth}}/\mathrm{NMI}_{\mathrm{real}}\). A positive signed gap indicates weaker measured dependence in the synthetic projection, whereas a negative signed gap indicates stronger measured dependence than in the corresponding real projection. The ratio is reported only for non-degenerate real projections with \(H(C_X)>0\), \(H(C_T)>0\), and \(\mathrm{NMI}_{\mathrm{real}}>0\). Because this ratio can be unstable when real dependence is close to the finite-sample shuffle floor, we interpret it together with the signed gap and the shuffle baseline. A ratio above one is therefore not automatically favorable.

For conditional diagnostics, we write \(\hat P_R\) and \(\hat P_S\) for the empirical real and synthetic projected distributions, respectively. We also compute conditional Jensen--Shannon divergence (cJSD) over text-cluster distributions given tabular states:
\begin{equation}
\begin{aligned}
\mathrm{cJSD}_{\mathcal U}
&=
\sum_{u\in\mathcal U} w_u d_u,\\
d_u
&=
\mathrm{JSD}\!\left(
\hat P_R(\cdot\mid u),
\hat P_S(\cdot\mid u)
\right),\\
w_u
&=
\frac{\hat P_R(C_X=u)}
{\sum_{u'\in\mathcal U}\hat P_R(C_X=u')}.
\end{aligned}
\label{eq:cjsd}
\end{equation}
Here \(\hat P_R(\cdot\mid u)\) and \(\hat P_S(\cdot\mid u)\) are the empirical real and synthetic text-cluster distributions conditional on \(C_X=u\). To reduce small-cell instability, cJSD is computed on \(\mathcal U=\{u:n_R(C_X=u)\ge n_{\min}^{R},\, n_S(C_X=u)\ge n_{\min}^{S}\}\), with \(n_{\min}^{R}=n_{\min}^{S}=5\). The threshold of 5 is a small-cell support rule: states with fewer records yield highly variable empirical conditional distributions and are therefore excluded from cJSD, while their excluded mass is reported through \(\mathrm{cov}_R\) and \(\mathrm{cov}_S\). The weights \(w_u\) renormalize real-data mass over \(\mathcal U\). We also report \(\mathrm{cov}_R=\sum_{u\in\mathcal U}\hat P_R(C_X=u)\) and \(\mathrm{cov}_S=\sum_{u\in\mathcal U}\hat P_S(C_X=u)\), so partial-support cJSD estimates are not compared as full-support estimates.

\subsection{Axis III: Utility via Cross-Modal Predictability}

Utility is evaluated using a train-on-synthetic, test-on-real (TSTR) predictive probe. In the text-to-attribute (T2A) direction used in our experiments, a dataset-specific classifier is trained on synthetic text embeddings to predict the corresponding tabular target and is evaluated on real holdout records. We report macro-averaged F1 for multiclass tasks. For the imbalanced Fake Jobs task, we additionally report balanced accuracy, the area under the receiver operating characteristic curve (AUROC), the area under the precision--recall curve (AUPRC), the Matthews correlation coefficient (MCC), and minority-class recall.

\subsection{Axis IV: Relative Joint-State Entropy}

To screen for projected cross-modal under-coverage or mode collapse, our framework measures diversity over the projected quantized table rather than over isolated marginals. For a synthetic projected distribution \(\hat Q\), we compute joint entropy and its real-relative version:
\begin{equation}
\begin{aligned}
H_{\hat Q}(C_X,C_T)
&= -\sum_{u,v}\hat q_{u,v}\log_2 \hat q_{u,v},\\
H_{\mathrm{rel}}(S)
&=
\frac{H_{\hat Q}(C_X,C_T)}
{H_{\hat P}(C_X,C_T)}.
\end{aligned}
\label{eq:hrel}
\end{equation}
where \(\hat q_{u,v}=\hat Q(C_X=u,C_T=v)\) and \(0\log 0=0\). A value substantially below one may indicate under-coverage, whereas a value above one may reflect either broader support or weakened dependence. We therefore interpret the entropy ratio jointly with fidelity and dependence metrics.

\subsection{Axis V: Holdout-Calibrated Record Proximity}

Record-proximity evaluation uses the distance to closest record (DCR) as an empirical nearest-neighbor diagnostic. For DCR only, tabular variables are transformed using feature-wise preprocessing fitted on the real training split, yielding \(X_{\mathrm{scaled}}\). Continuous variables are z-scored using real-training means and standard deviations, and continuous holdout and synthetic values outside the observed training range are clipped to \([\min,\max]\) before z-scoring. Categorical variables are mapped to integer codes using the training-split category vocabulary, with unseen synthetic categories mapped to an ``unknown'' code, and the resulting codes are z-scored using real-training means and standard deviations. This coding is used only to obtain a reproducible proximity screen; it is not a semantic ordinal model of categorical distance. Since nominal-category distances depend on the chosen encoding, the resulting DCR values and the holdout-calibrated proximity flag rate (PFR) should be interpreted as representation-level diagnostics rather than as privacy-risk estimates. Alternative mixed-type distances, such as one-hot or Gower-style encodings, are left to future work. Text embeddings are L2-normalized record-wise, yielding \(\tilde{E}\). The multimodal latent representation is
\begin{equation}
Z=[X_{\mathrm{scaled}}\parallel \lambda^\star \tilde E],
\label{eq:latent_z_revised}
\end{equation}
where \(\parallel\) denotes concatenation. Feature-wise tabular preprocessing avoids the scalar-feature degeneration caused by row-wise tabular normalization.

The scaling parameter \(\lambda^\star\) is estimated once per dataset using real training data:
\begin{equation}
\lambda^\star
=
\sqrt{
\frac{
\sum_r \mathrm{Var}(X_{\mathrm{scaled},r})
}{
\sum_s \mathrm{Var}(\tilde E_s)+\epsilon
}
},
\qquad
\epsilon=10^{-12}.
\label{eq:lambda}
\end{equation}
The same \(\lambda^\star\) is then applied to all synthetic methods for that dataset. Variances in Eq.~\ref{eq:lambda} are population variances computed over the real training split after the tabular preprocessing described above and after record-wise L2 normalization of text embeddings. This scaling equalizes the total empirical variance of the tabular and text blocks on the real training set. It does not guarantee equal influence on every nearest-neighbor comparison; we therefore report sensitivity to \(\lambda\).

For a synthetic record \(s\), DCR is
\begin{equation}
\mathrm{DCR}(s,R_{\mathrm{train}})
=
\min_{r\in R_{\mathrm{train}}}
\|Z_s-Z_r\|_2.
\label{eq:dcr}
\end{equation}
Rather than interpreting raw minimum DCR directly, we calibrate a proximity threshold using the real holdout-to-train DCR distribution. Let \(\tau_\alpha\) denote its \(\alpha\)-quantile:
\begin{equation}
\begin{aligned}
\tau_\alpha
&=
\operatorname{Quantile}_{\alpha}
\{d_h:h\in R_{\mathrm{holdout}}\},\\
d_h
&=
\mathrm{DCR}(h,R_{\mathrm{train}}).
\end{aligned}
\label{eq:tau}
\end{equation}
Because nearest-neighbor distances are computed numerically, we use a small numerical tolerance
\(\delta_{\mathrm{num}}=10^{-12}\) and define the effective calibrated threshold as
\begin{equation}
\tilde{\tau}_{\alpha}
=
\max(\tau_\alpha,\delta_{\mathrm{num}}).
\label{eq:tau_delta}
\end{equation}
The holdout-calibrated proximity flag rate is then
\begin{equation}
\mathrm{PFR}_{\alpha}(S)
=
\frac{1}{|S|}
\sum_{s\in S}
\mathbb{I}
\left[
\mathrm{DCR}(s,R_{\mathrm{train}})
\le
\tilde{\tau}_{\alpha}
\right].
\label{eq:pfr}
\end{equation}

We distinguish the dataset-level proximity flag rate from individual near-collision flags. The PFR value is the fraction of synthetic records that fall inside the holdout-calibrated proximity region under the selected representation and threshold. Separately, any synthetic record with \(\mathrm{DCR}\le 10^{-6}\) is treated as a near-collision requiring case-level inspection. A low PFR does not by itself imply privacy or high data quality, because unrealistic synthetic records may also lie far from the training data. Conversely, a high PFR indicates that many records are close under the selected representation, but it is not a formal estimate of disclosure probability. DCR and PFR are therefore reported as representation-level record-proximity diagnostics rather than as privacy guarantees.

\section{Experiments and Discussion}
\label{sec:experiments}

The proposed framework is a controlled multimodal evaluation rather than a single-score generator ranking. The experiments test whether common tabular-only and text-only metrics can pass deliberately misaligned records, and whether projection-level diagnostics reveal the resulting dependence, utility, diversity, and record-proximity changes. The real dataset \(R\) serves as the real-data reference. The text-permutation control constructs a negative control \(S_{\mathrm{tilt}}\) by shuffling text against tabular rows, preserving unimodal marginals while randomizing row-level cross-modal pairings.

Our experiments address four questions:
(\textbf{Q1}) Can conventional tabular-only and text-only metrics incorrectly validate misaligned multimodal records?
(\textbf{Q2}) Under what real-dependence conditions does the text-permutation control produce a detectable cross-modal violation?
(\textbf{Q3}) How do representative synthesis strategies preserve, destroy, or overstate cross-modal fidelity, dependence, diversity, utility, and record proximity?
(\textbf{Q4}) How sensitive are the conclusions to semantic resolution \(K\), DCR scaling \(\lambda/\lambda^\star\), class imbalance, sparse projected support, and higher-order dependence beyond pairwise projections?

\subsection{Experimental Setup}

\paragraph{Datasets.}
We evaluate on three real-world multimodal datasets:
(\emph{i}) \textbf{Amazon Reviews}~\cite{hou2024amazon}, containing review text paired with structured ratings;
(\emph{ii}) \textbf{Kiva Loans}~\cite{kiva2018kaggle}, containing borrower narratives paired with structured loan attributes such as sector; and
(\emph{iii}) \textbf{Fake Jobs, from the Employment Scam Aegean Dataset (EMSCAD)}~\cite{vidros2017automatic}, containing job-posting descriptions paired with fraud-related structured indicators.
After preprocessing, the retained evaluation subsets contain 6,064 Amazon records, 4,577 Kiva records, and 2,915 Fake Jobs records.

\paragraph{Baselines.}
We construct \(S_{\mathrm{tilt}}\) by shuffling the text column against original tabular rows. The text remains fluent and the tabular marginals remain unchanged, but row-level pairings are randomized. Its detectability depends on whether the evaluated projection contains measurable real cross-modal dependence. We compare \(S_{\mathrm{tilt}}\) against four learned baselines: \(S_{\mathrm{ind}}\), which generates tabular fields and text independently; \(S_{\mathrm{seq}}\), which generates tabular fields before conditioning text generation on them; \(S_{\mathrm{joint}}\), which generates both modalities autoregressively in a single prompt; and \(S_{\mathrm{lat}}\), our TabSyn-SBERT embedding-level adaptation, which treats SBERT text embeddings as continuous columns alongside the structured variables~\cite{zhang2024mixed}. TabSyn-SBERT does not generate surface text directly in this setup.

\paragraph{Implementation and data separation.}
The LLM baselines use an internal eBay deployment of GPT-5.2 with ten fixed in-context demonstrations, temperature~0.8, and up to three retries for malformed outputs. The released code uses GPT-4o-mini as a public substitute and includes the prompt-construction logic, retry policy, and raw artifacts. \(S_{\mathrm{ind}}\) uses CTGAN for the tabular fields and an independent LLM for text; \(S_{\mathrm{seq}}\) first generates the tabular fields and then generates text conditioned on them; \(S_{\mathrm{joint}}\) generates both modalities in one prompt; and \(S_{\mathrm{lat}}\) applies TabSyn-SBERT to structured variables and SBERT embeddings.

The fixed demonstration set \(D_{\mathrm{demo}}\) is disjoint from the real evaluation pool \(R\). Therefore, no demonstration record appears in either \(R_{\mathrm{train}}\) or \(R_{\mathrm{holdout}}\). We apply an 80/20 stratified split to \(R\). All data-dependent evaluation artifacts are fitted on \(R_{\mathrm{train}}\). The holdout partition is not used to fit these artifacts and is used for T2A testing and DCR-threshold calibration, while projection-level real reference distributions are computed on \(R\). Consequently, the reported LLM T2A scores do not involve direct prompt-example overlap with the real holdout set.

Because the CTGAN and TabSyn-SBERT outputs were produced before the post-hoc evaluation split, their T2A values and holdout-calibrated PFR values are interpreted as protocol-specific diagnostics rather than as strict estimates of generator generalization or privacy under a fully untouched holdout design.

The text-permutation control \(S_{\mathrm{tilt}}\) is constructed independently of the 10-shot prompting procedure by permuting text across real tabular rows. It is a deliberately constructed negative control rather than a learned generator. Its JSD, NMI, cJSD, and joint-entropy results test whether the evaluation metrics respond to disrupted pairings. Because it reuses real components by design, its T2A and PFR values are reported only as control responses, not as estimates of generator generalization or privacy.

\paragraph{Evaluation configuration.}
Text is encoded using Sentence-BERT (\texttt{all-MiniLM-L6-v2}, 384 dimensions)~\cite{reimers2019sentence}. The main setting uses \(K=20\), ten quantile bins for continuous variables, and 1{,}000 synthetic records per dataset--method pair. Text-cluster quantizers, tabular discretizers, categorical vocabularies, normalization statistics, and \(\lambda^\star\) are fitted on \(R_{\mathrm{train}}\) and applied unchanged to holdout and synthetic records; the SBERT encoder itself is pretrained and kept fixed.

T2A probes use frozen SBERT embeddings and prespecified classifier families with fixed hyperparameters: logistic regression for Fake Jobs (\texttt{class\_weight=balanced}, \texttt{max\_iter=1000}, \texttt{random\_state=42}) and gradient boosting for Amazon and Kiva (\texttt{n\_estimators=100}, \texttt{random\_state=42}; Kiva additionally uses \texttt{max\_depth=5}). We report tabular marginal fidelity, text similarity, JSD, NMI ratio and signed gap, cJSD, T2A, \(H_{\mathrm{rel}}\), and PFR; balanced utility metrics are additionally reported for Fake Jobs.

\subsection{Modality-Specific Metrics Do Not Test Cross-Modal Pairing}

Table~\ref{tab:traditional_metrics} reports the original modality-specific baseline scores. For methods that generate surface text, the reported BERTScore-F1 value compares each generated text with the reference text assigned by the evaluation pipeline. Because independently generated synthetic records do not have a natural one-to-one correspondence with real reference records, this index-matched score is interpreted only as a descriptive candidate--reference similarity score. It is not interpreted as a corpus-level text-distribution metric. TabSyn-SBERT does not generate surface text in this setup, so its embedding cosine similarity is reported in a separate column and is not compared directly with BERTScore-F1.

\begin{table}[t]
\centering
\scriptsize
\caption{Modality-specific baseline scores. Higher values indicate greater similarity within each individual metric column. BERTScore-F1 and embedding cosine use different inputs and scales and must not be compared across columns.}
\label{tab:traditional_metrics}
\renewcommand{\arraystretch}{1.10}
\setlength{\tabcolsep}{2.3pt}
\begin{tabularx}{\columnwidth}{@{}>{\raggedright\arraybackslash}p{0.14\columnwidth}>{\raggedright\arraybackslash}p{0.20\columnwidth}CCC@{}}
\toprule
Dataset &
Method &
\shortstack{Tabular\\marginal} &
\shortstack{BERTScore\\F1} &
\shortstack{Embedding\\cosine} \\
\midrule

\multirow{6}{*}{Amazon}
& \(R\)                  & 1.000 & 1.000 & 1.000 \\
& \(S_{\mathrm{tilt}}\)  & 1.000 & 0.840 & -- \\
& \(S_{\mathrm{ind}}\)   & 0.510 & 0.844 & -- \\
& \(S_{\mathrm{seq}}\)   & 0.510 & 0.852 & -- \\
& \(S_{\mathrm{joint}}\) & 0.635 & 0.846 & -- \\
& TabSyn-SBERT           & 0.794 & --    & 0.441 \\
\midrule

\multirow{6}{*}{Kiva}
& \(R\)                  & 1.000 & 1.000 & 1.000 \\
& \(S_{\mathrm{tilt}}\)  & 1.000 & 0.887 & -- \\
& \(S_{\mathrm{ind}}\)   & 0.590 & 0.889 & -- \\
& \(S_{\mathrm{seq}}\)   & 0.590 & 0.893 & -- \\
& \(S_{\mathrm{joint}}\) & 0.657 & 0.888 & -- \\
& TabSyn-SBERT           & 0.932 & --    & 0.755 \\
\midrule

\multirow{6}{*}{Fake Jobs}
& \(R\)                  & 1.000 & 1.000 & 1.000 \\
& \(S_{\mathrm{tilt}}\)  & 1.000 & 0.797 & -- \\
& \(S_{\mathrm{ind}}\)   & 0.815 & 0.809 & -- \\
& \(S_{\mathrm{seq}}\)   & 0.843 & 0.812 & -- \\
& \(S_{\mathrm{joint}}\) & 0.821 & 0.805 & -- \\
& TabSyn-SBERT           & 0.971 & --    & 0.553 \\
\bottomrule
\end{tabularx}

\par\vspace{0.45\baselineskip}
\parbox{\columnwidth}{\scriptsize
The tabular value is the originally reported marginal-fidelity score. BERTScore-F1 is reported only for methods that generate surface text. TabSyn-SBERT outputs SBERT embeddings rather than surface text, so its cosine similarity is included only as an embedding-level reference. No direct comparison is made between the BERTScore-F1 and embedding-cosine columns.}
\end{table}

The modality-specific scores do not test whether the text matches the structured attributes in the same record. As a corpus-level text-marginal control, we compute the JSD between the real and synthetic text-cluster marginals,
\(\mathrm{JSD}(\hat P_{C_T}\parallel\hat Q_{C_T})\),
while ignoring the tabular variables. The text-permutation control obtains 0.000 on all three datasets because it preserves the text corpus exactly. Its joint projected JSD is nevertheless nonzero on Amazon, Kiva, and Fake Jobs (0.102, 0.698, and 0.275, respectively). Thus, the text marginal can remain unchanged while the tabular--text pairing is disrupted. We therefore retain the index-matched BERTScore only as a descriptive modality-specific reference score and rely on the projected joint diagnostics for the cross-modal analysis.

\subsection{Main Projection-Level Results}

Before applying the permutation control, we compare each real projection with its shuffle baseline. Real/shuffled NMI is \(0.0103/0.0040\) for Amazon, \(0.6102/0.0098\) for Kiva, and \(0.0869/0.0032\) for Fake Jobs. Amazon is therefore reported separately as a weak-dependence boundary case; Kiva and Fake Jobs receive the complete multi-axis comparison.

\paragraph{Amazon: Weak-Dependence Boundary Case.}

\begin{table}[t]
\centering
\footnotesize
\caption{Projected JSD for the Amazon rating--text boundary case. Lower is closer to the real reference; weak real dependence limits the permutation control's discriminative power.}
\label{tab:amazon_boundary}
\renewcommand{\arraystretch}{1.08}
\setlength{\tabcolsep}{7pt}
\begin{tabular}{lc}
\toprule
Method & Projected JSD \\
\midrule
\(R\)                  & 0.0000 \\
\(S_{\mathrm{tilt}}\)  & 0.1017 \\
\(S_{\mathrm{ind}}\)   & 0.5778 \\
\(S_{\mathrm{seq}}\)   & 0.7369 \\
\(S_{\mathrm{joint}}\) & 0.5865 \\
TabSyn-SBERT           & 0.3219 \\
\bottomrule
\end{tabular}
\end{table}

Amazon is a weak-dependence boundary case: its real NMI (\(0.0103\)) is close to the permutation baseline (\(0.0040\)). Because \(S_{\mathrm{tilt}}\) preserves both modality marginals and the selected real dependence is weak, it attains lower projected JSD than the learned baselines. This does not validate the shuffled records; it shows that the selected projection has limited discriminative power. Its high PFR instead reflects deliberate reuse of real components and should be interpreted separately from cross-modal alignment.

\paragraph{Kiva and Fake Jobs: Complete Multi-Axis Comparison.}

Table~\ref{tab:axis_summary} reports all metrics for Kiva and Fake Jobs. The metrics are complementary and are not combined into an overall ranking. The row \(R\) anchors the distributional columns; T2A and PFR are not applicable to it.

\begin{table*}[!t]
\centering
\scriptsize
\caption{Projection-level results for Kiva and Fake Jobs. Lower JSD/cJSD and smaller \(\lvert\Delta_{\mathrm{NMI}}\rvert\) are preferred; higher T2A is better; \(H_{\mathrm{rel}}\) is interpreted relative to 1. PFR is a proximity flag rate, not a quality or privacy score. NMI is ratio / signed gap. Bold marks best directly comparable synthetic values; partial-coverage cJSD is excluded.}
\label{tab:axis_summary}
\renewcommand{\arraystretch}{1.08}
\setlength{\tabcolsep}{3.2pt}
\begin{tabularx}{\textwidth}{@{}llCCCCCC@{}}
\toprule
Dataset &
Method &
Proj. JSD \(\downarrow\) &
NMI ratio / \(\Delta_{\mathrm{NMI}}\) &
cJSD \(\downarrow\) &
T2A \(\uparrow\) &
\(H_{\mathrm{rel}}\) &
PFR$_{1\%}$ \\
\midrule

\multirow{6}{*}{Kiva}
& \(R\)                  & 0.000          & 1.00 / 0.000         & 0.000                     & --             & 1.000          & --    \\
& \(S_{\mathrm{tilt}}\)  & 0.698          & 0.016 / 0.600        & 0.687                     & 0.076          & 1.432          & 0.186 \\
& \(S_{\mathrm{ind}}\)   & 0.844          & 0.075 / 0.565        & 0.773                     & 0.062          & 1.601          & 0.000 \\
& TabSyn-SBERT           & 0.727          & 0.791 / 0.128        & 0.363$^\dagger$           & \textbf{0.427} & 1.093          & 0.000 \\
& \(S_{\mathrm{seq}}\)   & \textbf{0.602} & \textbf{0.948 / 0.032} & \textbf{0.448}$^\ddagger$ & 0.345        & 1.120          & 0.010 \\
& \(S_{\mathrm{joint}}\) & 0.729          & 1.100 / -0.061       & 0.623                     & 0.331          & \textbf{0.957} & 0.000 \\
\midrule

\multirow{6}{*}{Fake Jobs}
& \(R\)                  & 0.000          & 1.00 / 0.000         & 0.000          & --             & 1.000          & --    \\
& \(S_{\mathrm{tilt}}\)  & 0.275          & 0.037 / 0.084        & 0.215          & 0.215          & 0.867          & 0.580 \\
& \(S_{\mathrm{ind}}\)   & 0.458          & 0.052 / 0.082        & 0.352          & 0.372          & \textbf{0.878} & 0.000 \\
& TabSyn-SBERT           & \textbf{0.165} & \textbf{0.848 / 0.013} & \textbf{0.126} & \textbf{0.555} & 0.836       & 0.000 \\
& \(S_{\mathrm{seq}}\)   & 0.522          & 2.319 / -0.115       & 0.433          & 0.343          & 0.819          & 0.000 \\
& \(S_{\mathrm{joint}}\) & 0.568          & 2.556 / -0.135       & 0.443          & 0.341          & 0.833          & 0.000 \\
\bottomrule
\end{tabularx}

\par\vspace{0.45\baselineskip}
\parbox{\textwidth}{\scriptsize
\(^{\dagger}\) TabSyn-SBERT omits 5 of 13 Kiva sectors (\(\mathrm{cov}_R=0.67\), \(\mathrm{cov}_S=1.00\)); its cJSD is partial-coverage.
\(^{\ddagger}\) Best full-coverage Kiva cJSD.}
\end{table*}

Table~\ref{tab:axis_summary} is a diagnostic table rather than a leaderboard. Its axes describe different properties of the synthetic data and should not be collapsed into a single overall ranking. In particular, a method can preserve the measured dependence level while failing to cover important tabular states, or it can achieve a low PFR simply by generating records that are far from the real-data distribution.

On Kiva, \(S_{\mathrm{tilt}}\) and \(S_{\mathrm{ind}}\) largely remove the real sector--text dependence. Among the learned methods with full conditional-state coverage, \(S_{\mathrm{seq}}\) is closest to the real NMI and has the lowest full-coverage cJSD. \(S_{\mathrm{joint}}\) exhibits slightly stronger measured dependence than the real projection. TabSyn-SBERT obtains the highest T2A score and a lower numerical cJSD, but its cJSD applies only to the sectors that it generates; several real sectors receive no synthetic records. Its conditional result must therefore be interpreted together with the reported coverage rather than as uniformly better alignment.

On Fake Jobs, TabSyn-SBERT has the lowest projected JSD and cJSD, the smallest absolute NMI gap, and the highest T2A score. By contrast, \(S_{\mathrm{seq}}\) and \(S_{\mathrm{joint}}\) exhibit measured fraud--text dependence more than twice the real-data level, while \(S_{\mathrm{tilt}}\) and \(S_{\mathrm{ind}}\) retain only a small fraction of the real dependence. These results show that cross-modal fidelity requires a measured dependence level close to the real-data reference, rather than either minimal dependence or maximally strong dependence. Because TabSyn-SBERT directly models the same SBERT representation used by the text clusters and predictive probe, its results should be interpreted as strong embedding-level performance under the present protocol, not as evidence of superior surface-text generation or a general advantage of diffusion models over LLMs.

\subsection{Cross-Modal Utility under Class Imbalance}

The T2A column in Table~\ref{tab:axis_summary} reports the main utility metric. For Fake Jobs, the target is imbalanced, so positive-class F1 alone can be misleading. Table~\ref{tab:fakejobs_balanced} reports balanced metrics.

\begin{table}[t]
\centering
\footnotesize
\caption{Fake Jobs T2A utility under class imbalance. Higher is better for all reported utility metrics. Bold values mark the best value in each metric column among non-reference methods; ties are bolded. Min. Recall denotes minority-class recall. Values are rounded; ``--'' denotes a score-based metric that was not computed for the hard-label majority baseline.}
\label{tab:fakejobs_balanced}
\renewcommand{\arraystretch}{1.1}
\resizebox{\columnwidth}{!}{%
\begin{tabular}{lcccccc}
\toprule
Method & Pos. F1 & Balanced Acc. & AUROC & AUPRC & MCC & Min. Recall \\
\midrule
Majority & 0.000 & 0.500 & -- & -- & -- & 0.000 \\
\(S_{\mathrm{ind}}\) & 0.372 & 0.604 & 0.604 & 0.286 & 0.166 & \textbf{0.741} \\
\(S_{\mathrm{seq}}\) & 0.343 & 0.598 & 0.622 & 0.401 & 0.239 & 0.277 \\
\(S_{\mathrm{joint}}\) & 0.341 & 0.599 & 0.611 & 0.314 & 0.260 & 0.259 \\
TabSyn-SBERT & \textbf{0.555} & \textbf{0.760} & \textbf{0.845} & \textbf{0.679} & \textbf{0.439} & \textbf{0.741} \\
\(S_{\mathrm{tilt}}\) & 0.215 & 0.462 & 0.463 & 0.185 & -0.061 & 0.321 \\
\bottomrule
\end{tabular}%
}
\end{table}

The balanced metrics confirm that the Fake Jobs result is not an artifact of positive-class F1 alone. The text-permutation control has balanced accuracy below 0.5, AUROC close to random, low AUPRC, and negative MCC. Under the fixed SBERT-based probe, TabSyn-SBERT achieves the strongest utility scores and the closest measured dependence level among the evaluated methods. Because TabSyn-SBERT directly models the same SBERT representation used by the text clusters and the predictive probe, this is a representation-matched comparison. It should not be interpreted as evidence of superior surface-text generation or as a general advantage of diffusion models over LLMs.

\subsection{Holdout-Calibrated Record-Proximity Analysis}

Table~\ref{tab:privacy_extended} reports holdout-calibrated DCR thresholds and PFR. Thresholds are displayed to three decimals, but PFR is computed before rounding using \(\tilde{\tau}_{\alpha}\); hence 0.000 denotes numerical zero or below the display precision in the DCR representation, not byte-identical raw records.

\begin{table}[!t]
\centering
\scriptsize
\caption{Holdout-calibrated DCR thresholds and PFR values. Thresholds are shown to three decimals, but PFR is computed before rounding using \(\tilde{\tau}_{\alpha}=\max(\tau_\alpha,\delta_{\mathrm{num}})\). For Kiva and Fake Jobs, the displayed \(\tau_{5\%}\) and \(\tau_{10\%}\) remain 0.000 under the selected DCR representation, so PFR\(_{1\%}\), PFR\(_{5\%}\), and PFR\(_{10\%}\) coincide.}
\label{tab:privacy_extended}
\renewcommand{\arraystretch}{1.08}
\setlength{\tabcolsep}{2pt}
\begin{tabularx}{\columnwidth}{@{}lccccCC@{}}
\toprule
Dataset & \(\lambda^\star\) & \(\tau_{1\%}\) & \(\tau_{5\%}\) & \(\tau_{10\%}\) & Max learned PFR & Tilted PFR \\
\midrule
Amazon 
& 1.138 & 0.000 & 0.553 & 0.643 
& 0.000 / 0.014 / 0.032 
& 0.388 / 0.490 / 0.508 \\
Kiva 
& 1.315 & 0.000 & 0.000 & 0.000 
& 0.010 / 0.010 / 0.010 
& 0.186 / 0.186 / 0.186 \\
Fake Jobs 
& 1.165 & 0.000 & 0.000 & 0.000 
& 0.000 / 0.000 / 0.000 
& 0.580 / 0.580 / 0.580 \\
\bottomrule
\end{tabularx}

\par\vspace{0.45\baselineskip}
\noindent\begin{minipage}{\columnwidth}
\scriptsize
Learned PFR is the maximum over \(S_{\mathrm{ind}}\), \(S_{\mathrm{seq}}\), \(S_{\mathrm{joint}}\), and TabSyn-SBERT, reported as PFR\(_{1\%}\) / PFR\(_{5\%}\) / PFR\(_{10\%}\). Collision and near-collision statements refer to the DCR representation after preprocessing and do not necessarily imply byte-identical raw records. We use \(\mathrm{DCR}\le 10^{-6}\) as an individual near-collision inspection threshold; dataset-level conclusions are based on the calibrated PFR rather than raw minimum DCR alone.
\end{minipage}
\end{table}

The calibrated DCR diagnostic flags the learned baselines at low rates, whereas \(S_{\mathrm{tilt}}\) has a high PFR because it reuses real tabular and text components and many of its records fall inside the calibrated proximity region. We distinguish dataset-level PFR from individual near-collision flags: records with \(\mathrm{DCR}\le 10^{-6}\) require case-level inspection, whereas dataset-level conclusions are based on PFR. Under this threshold, no learned baseline produces a near-collision on Amazon. On Kiva, \(S_{\mathrm{seq}}\) produces approximately ten numerical-zero records (\(\mathrm{PFR}_{1\%}=0.010\)), indicating representation-level duplicates rather than broadly elevated record proximity. For Kiva and Fake Jobs, the displayed calibration threshold remains 0.000 through the 10th percentile; this reflects redundancy in the evaluated representation and is not evidence of a formal privacy breach.

\subsection{Sensitivity and Failure Analysis}

Table~\ref{tab:robustness_summary} summarizes robustness and boundary-condition checks. These checks probe three practical choices that affect interpretation: the semantic resolution \(K\), the DCR scaling parameter \(\lambda^\star\), and the ability of pairwise projections to detect higher-order interactions.

\begin{table}[!t]
\centering
\scriptsize
\caption{Robustness and boundary-condition summary.}
\label{tab:robustness_summary}
\renewcommand{\arraystretch}{1.08}
\setlength{\tabcolsep}{2.5pt}
\begin{tabularx}{\columnwidth}{@{}p{0.23\columnwidth}p{0.25\columnwidth}X@{}}
\toprule
Check & Setting & Main finding \\
\midrule
\(K\)-sensitivity 
& \(K\in\{5,10,15,20,30\}\) 
& Fake Jobs remains stable, with tilted-to-TabSyn-SBERT JSD ratios from 2.54 to 1.49. Amazon remains weak because rating--text dependence is small. Kiva is sparse under the primary sector projection, with occupied-cell ratios 20--51\% and minimum cell count 1. \\
\midrule
\(\lambda\)-sensitivity 
& \(\lambda/\lambda^\star\in\{0.25,0.5,1,2,4\}\) 
& Raw DCR changes with \(\lambda\), but PFR-based flagging pattern is unchanged across this 16-fold range: learned baselines remain unflagged or nearly unflagged, while \(S_{\mathrm{tilt}}\) remains flagged. \\
\midrule
XOR/parity stress test 
& \(X_1,X_2,X_3\in\{0,1\}\), \(C_T=X_1\oplus X_2\oplus X_3\); \(S_{\mathrm{tilt}}\) shuffles \(C_T\) while preserving the tabular marginal 
& Each pair \((X_j,C_T)\) is marginally independent, so the mean pairwise JSD remains near zero in the finite-sample evaluation (0.0075). Shuffling destroys the parity support constraint, and the targeted 3-way projection detects the violation with empirical JSD 0.5542. This value is a finite-sample diagnostic statistic rather than the closed-form JSD of an ideal uniform parity distribution. \\
\bottomrule
\end{tabularx}
\end{table}

For interpretation, we flag a projection as sparse when its occupied-cell ratio is at most 30\% or its minimum synthetic conditional support is below \(n_{\min}^{S}=5\). Under these checks, Fake Jobs is stable at \(K=20\), Amazon is marginal, and Kiva is sparse under the primary sector projection. The XOR/parity stress test further shows that pairwise projections can miss higher-order dependence by construction. Pairwise projections should therefore be treated as scalable diagnostics rather than full joint-distribution estimators.

\subsection{Limitations}

The framework is a projection-based evaluation, not a full estimator of the continuous multimodal joint distribution. The framework is most informative when the selected projection contains real dependence clearly separated from the empirical shuffle baseline. Amazon illustrates the opposite boundary case, in which weak measured dependence limits the discriminative power of permutation-based diagnostics. JSD can reflect marginal mismatch as well as dependence mismatch; NMI ratios can be unstable near the shuffle floor and can exceed one under stronger measured dependence than in the real-data projection; and high \(H_{\mathrm{rel}}\) does not imply semantic validity. Pairwise projections can miss higher-order interactions, sparse cJSD estimates require coverage reporting, and DCR/PFR are empirical proximity diagnostics rather than formal privacy guarantees. Metric magnitudes should be compared primarily within the same dataset, projection, and semantic resolution \(K\). Preserving the real-data dependence structure in selected projections is a necessary dataset-level criterion, but it is not sufficient evidence of instance-level semantic consistency.

\section{Conclusion}
\label{sec:conclusion}

This paper shows that favorable modality-specific scores are not sufficient to validate the pairing between structured attributes and text in a synthetic dataset. We extend SynEval with fixed text-embedding quantization, selected tabular--text projections, a text-permutation control, and complementary diagnostics for projected fidelity, measured dependence, predictive utility, joint-state entropy, and representation-level record proximity.

Across Amazon Reviews, Kiva Loans, and Fake Jobs, the results show that the usefulness of a projection depends on the level of real-data dependence it contains. The Kiva and Fake Jobs projections are clearly separated from their shuffle baselines and reveal substantial changes after text permutation or independent generation. Amazon instead provides a weak-dependence boundary case in which the selected rating--text projection has limited power to distinguish disrupted pairings. Some conditioned LLM baselines also exhibit stronger measured dependence than the corresponding real-data projections, while TabSyn-SBERT performs strongly under the embedding-level evaluation protocol.

These findings support explicit comparison of synthetic and real cross-modal association rather than reliance on unimodal scores or on high dependence alone. The framework remains a projection-based dataset-level evaluator: preserving the measured dependence structure in selected projections is a necessary criterion for cross-modal fidelity, but it is not sufficient evidence that every individual record is semantically correct. Sparse projected support, weak dependence near the shuffle floor, and higher-order interactions require separate reporting, while DCR and PFR should be interpreted only as representation-level record-proximity diagnostics.

\section*{Acknowledgment}
This work was supported in part by a research grant from eBay. The authors thank eBay for its support of this research. The views and conclusions expressed in this paper are those of the authors and do not necessarily reflect the official policies or positions of eBay.

\section*{Artifact Availability}
The source code, datasets, and comprehensive instructions required to reproduce the evaluation in this paper are publicly available on GitHub at \url{https://github.com/privacy-enhancing-technologies/SynEval}. The evaluated and permanently archived version of this artifact is available at \href{https://zenodo.org/records/21786626}{DOI: 10.5281/zenodo.21786626}.

\section*{Declaration of Generative AI Use}
Generative AI systems were used in two ways. First, an internal eBay deployment of GPT-5.2 was used as an experimental generative baseline, as described in Section~\ref{sec:experiments}. Second, generative AI tools were used to assist with scripting, automation, and grammar checking. All AI-assisted code, generated artifacts, and experimental analyses were checked by the authors before inclusion in the paper.

\FloatBarrier
\bibliographystyle{IEEEtran}
\bibliography{references}

@article{heckerman2008tutorial,
  title={A tutorial on learning with Bayesian networks},
  author={Heckerman, David},
  journal={Innovations in Bayesian networks: Theory and applications},
  pages={33--82},
  year={2008},
  publisher={Springer}
}

@article{frees1998understanding,
  title={Understanding relationships using copulas},
  author={Frees, Edward W and Valdez, Emiliano A},
  journal={North American actuarial journal},
  volume={2},
  number={1},
  pages={1--25},
  year={1998},
  publisher={Taylor \& Francis}
}

@article{xu2019modeling,
  title={Modeling tabular data using conditional gan},
  author={Xu, Lei and Skoularidou, Maria and Cuesta-Infante, Alfredo and Veeramachaneni, Kalyan},
  journal={Advances in neural information processing systems},
  volume={32},
  year={2019}
}

@article{touvron2023llama,
  title={Llama 2: Open foundation and fine-tuned chat models},
  author={Touvron, Hugo and Martin, Louis and Stone, Kevin and Albert, Peter and Almahairi, Amjad and Babaei, Yasmine and Bashlykov, Nikolay and Batra, Soumya and Bhargava, Prajjwal and Bhosale, Shruti and others},
  journal={arXiv preprint arXiv:2307.09288},
  year={2023}
}

@article{borisov2022language,
  title={Language models are realistic tabular data generators},
  author={Borisov, Vadim and Se{\ss}ler, Kathrin and Leemann, Tobias and Pawelczyk, Martin and Kasneci, Gjergji},
  journal={arXiv preprint arXiv:2210.06280},
  year={2022}
}

@inproceedings{kotelnikov2023tabddpm,
  title={Tabddpm: Modelling tabular data with diffusion models},
  author={Kotelnikov, Akim and Baranchuk, Dmitry and Rubachev, Ivan and Babenko, Artem},
  booktitle={International conference on machine learning},
  pages={17564--17579},
  year={2023},
  organization={PMLR}
}

@article{zhou2025missddim,
  title={MissDDIM: Deterministic and Efficient Conditional Diffusion for Tabular Data Imputation},
  author={Zhou, Youran and Bouadjenek, Mohamed Reda and Aryal, Sunil},
  journal={arXiv preprint arXiv:2508.03083},
  year={2025}
}

@article{villaizan2025diffusion,
  title={Diffusion models for tabular data imputation and synthetic data generation},
  author={Villaiz{\'a}n-Vallelado, Mario and Salvatori, Matteo and Segura, Carlos and Arapakis, Ioannis},
  journal={ACM Transactions on Knowledge Discovery from Data},
  volume={19},
  number={6},
  pages={1--32},
  year={2025},
  publisher={ACM New York, NY}
}

@article{brown2020language,
  title={Language models are few-shot learners},
  author={Brown, Tom and Mann, Benjamin and Ryder, Nick and Subbiah, Melanie and Kaplan, Jared D and Dhariwal, Prafulla and Neelakantan, Arvind and Shyam, Pranav and Sastry, Girish and Askell, Amanda and others},
  journal={Advances in neural information processing systems},
  volume={33},
  pages={1877--1901},
  year={2020}
}

@article{kim2024epic,
  title={Epic: Effective prompting for imbalanced-class data synthesis in tabular data classification via large language models},
  author={Kim, Jinhee and Kim, Taesung and Choo, Jaegul},
  journal={Advances in Neural Information Processing Systems},
  volume={37},
  pages={31504--31542},
  year={2024}
}

@article{zhao2023tabula,
  title={Tabula: Harnessing language models for tabular data synthesis},
  author={Zhao, Zilong and Birke, Robert and Chen, Lydia},
  journal={arXiv preprint arXiv:2310.12746},
  year={2023}
}

@techreport{wef2025synthetic,
  author      = {{World Economic Forum}},
  title       = {Synthetic Data: The New Data Frontier},
  institution = {World Economic Forum},
  type        = {Briefing Paper},
  year        = {2025},
  month       = sep,
  url         = {https://reports.weforum.org/docs/WEF_Synthetic_Data_2025.pdf},
  note        = {Accessed: 2026-07-03}
}

@article{huang2019clinicalbert,
  title={ClinicalBERT: Modeling clinical notes and predicting hospital readmission},
  author={Huang, Kexin and Altosaar, Jaan and Ranganath, Rajesh},
  journal={arXiv preprint arXiv:1904.05342},
  year={2019}
}

@book{silverman2018density,
  title={Density estimation for statistics and data analysis},
  author={Silverman, Bernard W},
  year={2018},
  publisher={Routledge}
}

@book{koller2009probabilistic,
  title={Probabilistic graphical models: principles and techniques},
  author={Koller, Daphne and Friedman, Nir},
  year={2009},
  publisher={MIT press}
}

@article{cooper1990computational,
  title={The computational complexity of probabilistic inference using Bayesian belief networks},
  author={Cooper, Gregory F},
  journal={Artificial intelligence},
  volume={42},
  number={2-3},
  pages={393--405},
  year={1990},
  publisher={Elsevier}
}

@misc{gdpr2016,
  title        = {Regulation ({EU}) 2016/679 of the European Parliament and of the Council of 27 April 2016 on the protection of natural persons with regard to the processing of personal data and on the free movement of such data, and repealing Directive 95/46/EC (General Data Protection Regulation)},
  author       = {{European Union}},
  year         = {2016},
  journal      = {Official Journal of the European Union},
  volume       = {L 119},
  pages        = {1--88},
  url          = {https://eur-lex.europa.eu/eli/reg/2016/679/oj}
}

@misc{ccpa2018,
  title        = {California Consumer Privacy Act of 2018 ({CCPA})},
  author       = {{California State Legislature}},
  year         = {2018},
  note         = {California Civil Code \S\S 1798.100 et seq.},
  url          = {https://leginfo.legislature.ca.gov/faces/billTextClient.xhtml?bill_id=201720180AB375}
}

@inproceedings{hegselmann2023tabllm,
  title={TabLLM: Few-shot classification of tabular data with large language models},
  author={Hegselmann, Stefan and Buendia, Alejandro and Lang, Hunter and Agrawal, Monica and Jiang, Xiaoyi and Sontag, David},
  booktitle={International Conference on Artificial Intelligence and Statistics},
  pages={5549--5581},
  year={2023},
  organization={PMLR}
}

@article{fang2024large,
  title={Large language models on tabular data: A survey},
  author={Fang, Xi and Xu, Weijie and Tan, Fiona Anting and Zhang, Jiani and Hu, Ziqing and Qi, Yanjun and Nickleach, Scott and Socolinsky, Diego and Sengamedu, Srinivasan and Faloutsos, Christos},
  journal={arXiv preprint arXiv:2402.17944},
  year={2024}
}

@article{baltrusaitis2018multimodal,
  title={Multimodal machine learning: A survey and taxonomy},
  author={Baltru{\v{s}}aitis, Tadas and Ahuja, Chaitanya and Morency, Louis-Philippe},
  journal={IEEE transactions on pattern analysis and machine intelligence},
  volume={41},
  number={2},
  pages={423--443},
  year={2018},
  publisher={IEEE}
}

@article{gaw2022multimodal,
  title={Multimodal data fusion for systems improvement: A review},
  author={Gaw, Nathan and Yousefi, Safoora and Gahrooei, Mostafa Reisi},
  journal={IISE Transactions},
  volume={54},
  number={11},
  pages={1098--1116},
  year={2022},
  publisher={Taylor \& Francis}
}

@inproceedings{patki2016synthetic,
  title={The synthetic data vault},
  author={Patki, Neha and Wedge, Roy and Veeramachaneni, Kalyan},
  booktitle={2016 IEEE international conference on data science and advanced analytics (DSAA)},
  pages={399--410},
  year={2016},
  organization={IEEE}
}

@inproceedings{papineni2002bleu,
  title={Bleu: a method for automatic evaluation of machine translation},
  author={Papineni, Kishore and Roukos, Salim and Ward, Todd and Zhu, Wei-Jing},
  booktitle={Proceedings of the 40th annual meeting of the Association for Computational Linguistics},
  pages={311--318},
  year={2002}
}

@inproceedings{lin2004rouge,
  title={Rouge: A package for automatic evaluation of summaries},
  author={Lin, Chin-Yew},
  booktitle={Text summarization branches out},
  pages={74--81},
  year={2004}
}

@article{zhang2019bertscore,
  title={Bertscore: Evaluating text generation with bert},
  author={Zhang, Tianyi and Kishore, Varsha and Wu, Felix and Weinberger, Kilian Q and Artzi, Yoav},
  journal={arXiv preprint arXiv:1904.09675},
  year={2019}
}

@inproceedings{pillutla2021mauve,
  title={MAUVE: Measuring the Gap Between Neural Text and Human Text using Divergence Frontiers},
  author={Pillutla, Krishna and Swayamdipta, Swabha and Zellers, Rowan and Thickstun, John and Welleck, Sean and Choi, Yejin and Harchaoui, Zaid},
  booktitle={Advances in Neural Information Processing Systems (NeurIPS)},
  volume={34},
  pages={4816--4828},
  year={2021}
}

@article{qian2023synthcity,
  title={Synthcity: facilitating innovative use cases of synthetic data in different data modalities},
  author={Qian, Zhaozhi and Cebere, Bogdan-Constantin and van der Schaar, Mihaela},
  journal={arXiv preprint arXiv:2301.07573},
  year={2023}
}

@inproceedings{choi2017generating,
  title={Generating multi-label discrete patient records using generative adversarial networks},
  author={Choi, Edward and Biswal, Siddharth and Malin, Bradley and Duke, Jon and Stewart, Walter F and Sun, Jimeng},
  booktitle={Machine learning for healthcare conference},
  pages={286--305},
  year={2017},
  organization={PMLR}
}

@inproceedings{aggarwal2001surprising,
  title={On the surprising behavior of distance metrics in high dimensional space},
  author={Aggarwal, Charu C and Hinneburg, Alexander and Keim, Daniel A},
  booktitle={International conference on database theory},
  pages={420--434},
  year={2001},
  organization={Springer}
}

@inproceedings{krause2006near,
  title={Near-optimal sensor placements: Maximizing information while minimizing communication cost},
  author={Krause, Andreas and Guestrin, Carlos and Gupta, Anupam and Kleinberg, Jon},
  booktitle={Proceedings of the 5th international conference on Information processing in sensor networks},
  pages={2--10},
  year={2006}
}

@inproceedings{reimers2019sentence,
  title={Sentence-BERT: Sentence Embeddings using Siamese BERT-Networks},
  author={Reimers, Nils and Gurevych, Iryna},
  booktitle={Proceedings of the 2019 Conference on Empirical Methods in Natural Language Processing and the 9th International Joint Conference on Natural Language Processing (EMNLP-IJCNLP)},
  pages={3982--3992},
  year={2019}
}

@book{cover2006elements,
  title={Elements of Information Theory},
  author={Cover, Thomas M and Thomas, Joy A},
  year={2006},
  publisher={John Wiley \& Sons}
}

@article{johnson2016mimic,
  title={MIMIC-III, a freely accessible critical care database},
  author={Johnson, Alistair EW and Pollard, Tom J and Shen, Lu and Lehman, Li-wei H and Feng, Mengling and Ghassemi, Mohammad and Moody, Benjamin and Szolovits, Peter and Anthony, Leo and Mark, Roger G},
  journal={Scientific data},
  volume={3},
  number={1},
  pages={1--9},
  year={2016},
  publisher={Nature Publishing Group}
}

@misc{kiva2018kaggle,
  title={Data Science for Good: Kiva Crowdfunding},
  author={Kaggle and Kiva},
  year={2018},
  howpublished={\url{https://www.kaggle.com/datasets/kiva/data-science-for-good-kiva-crowdfunding}},
  note={Accessed: 2024}
}

@article{vidros2017automatic,
  title={Automatic detection of online recruitment frauds: Characteristics, methods, and a public dataset},
  author={Vidros, Sokratis and Kolias, Constantinos and Kambourakis, Georgios and Akoglu, Leman},
  journal={Future Internet},
  volume={9},
  number={1},
  pages={6},
  year={2017},
  publisher={MDPI}
}

@article{hou2024amazon,
  title={Bridging Language and Items for Retrieval and Recommendation},
  author={Hou, Yupeng and Li, Jiacheng and He, Zhankui and Yan, An and Chen, Xiusi and McAuley, Julian},
  journal={arXiv preprint arXiv:2403.03952},
  year={2024}
}

@inproceedings{zhang2024mixed,
  title={Mixed-type tabular data synthesis with score-based diffusion in latent space},
  author={Zhang, Hengrui and Zhang, Jiani and Shen, Zhengyuan and Srinivasan, Balasubramaniam and Qin, Xiao and Faloutsos, Christos and Rangwala, Huzefa and Karypis, George},
  booktitle={International Conference on Learning Representations},
  volume={2024},
  pages={52829--52857},
  year={2024}
}

@article{yuan2024multi,
  title={A multi-faceted evaluation framework for assessing synthetic data generated by large language models},
  author={Yuan, Yefeng and Liu, Yuhong and Cheng, Liang},
  journal={arXiv preprint arXiv:2404.14445},
  year={2024}
}
\end{document}